\documentclass[runningheads]{llncs}
\usepackage{graphicx}
\usepackage{amsmath,amssymb} % define this before the line numbering.
\usepackage{color}
\usepackage[width=122mm,left=12mm,paperwidth=146mm,height=193mm,top=12mm,paperheight=217mm]{geometry}

\usepackage{subcaption}
\usepackage{adjustbox}
\usepackage{amsmath}
\usepackage{amssymb}
\usepackage{array}
\usepackage{bm}
\usepackage{booktabs}
\usepackage{color}
\usepackage{epsfig}
\usepackage{graphicx}
\usepackage{multirow}
\usepackage{url}
\usepackage{adjustbox}
\usepackage{xspace}
\usepackage[pagebackref=true,breaklinks=true,colorlinks,bookmarks=false]{hyperref}

\usepackage{array}
\newcolumntype{L}[1]{>{\raggedright\let\newline\\\arraybackslash\hspace{0pt}}m{#1}}
\newcolumntype{C}[1]{>{\centering\let\newline\\\arraybackslash\hspace{0pt}}m{#1}}
\newcolumntype{R}[1]{>{\raggedleft\let\newline\\\arraybackslash\hspace{0pt}}m{#1}}

\makeatletter
\DeclareRobustCommand\onedot{\futurelet\@let@token\@onedot}
\def\@onedot{\ifx\@let@token.\else.\null\fi\xspace}

\def\eg{\emph{e.g}\onedot}

\def\etc{\emph{etc}\onedot} 
 
\def\etal{\emph{et al}\onedot}

\def\cX{\mathcal{X}}
\def\cY{\mathcal{Y}}

\def\bR{\mathbb{R}}

\def\datasetName{Keypoint-5\xspace}
\def\model{3D INterpreter Network\xspace}
\def\modelshort{3D-INN\xspace}
\makeatother

\newcommand{\tb}[1]{\textbf{#1}}
\newcommand{\sect}[1]{Section~\ref{#1}}
\newcommand{\fig}[1]{Figure~\ref{#1}}

\newcommand{\xpar}[1]{\noindent\textbf{#1}\ \ }
\newcommand{\vpar}[1]{\vspace{1.5mm}\noindent\textbf{#1}\ \ }

\definecolor{MyDarkBlue}{rgb}{0,0.08,1}
\definecolor{MyDarkGreen}{rgb}{0.02,0.6,0.02}
\definecolor{MyDarkRed}{rgb}{0.8,0.02,0.02}
\definecolor{MyDarkOrange}{rgb}{0.40,0.2,0.02}
\definecolor{MyPurple}{RGB}{111,0,255}
\definecolor{MyRed}{rgb}{1.0,0.0,0.0}
\definecolor{MyGold}{rgb}{0.75,0.6,0.12}
\definecolor{MyDarkgray}{rgb}{0.66, 0.66, 0.66}

\begin{document}

\pagestyle{headings}
\mainmatter

\title{Single Image 3D Interpreter Network}

\author{
Jiajun Wu$^{1*}$, Tianfan Xue$^{1*}$, Joseph J. Lim$^{1,2}$, Yuandong Tian$^3$,\\ Joshua B. Tenenbaum$^1$, Antonio Torralba$^1$, and William T. Freeman$^{1,4}$}
\institute{
$^1$Massachusetts Institute of Technology\\
$^2$Stanford University\qquad
$^3$Facebook AI Research\qquad
$^4$Google Research
}

\titlerunning{Single Image 3D Interpreter Network}
\authorrunning{J. Wu et al.}

\maketitle
\footnotetext{$*$ indicates equal contributions.}

\begin{abstract}

Understanding 3D object structure from a single image is an important but difficult task in computer vision, mostly due to the lack of 3D object annotations in real images. Previous work tackles this problem by either solving an optimization task given 2D keypoint positions, or training on synthetic data with ground truth 3D information.

\setlength{\parindent}{15pt}
In this work, we propose \model (\modelshort), an end-to-end framework which sequentially estimates 2D keypoint heatmaps and 3D object structure, trained on both real 2D-annotated images and synthetic 3D data. This is made possible mainly by two technical innovations. First, we propose a Projection Layer, which projects estimated 3D structure to 2D space, so that \modelshort can be trained to predict 3D structural parameters supervised by 2D annotations on real images. Second, heatmaps of keypoints serve as an intermediate representation connecting real and synthetic data, enabling \modelshort to benefit from the variation and abundance of synthetic 3D objects, without suffering from the difference between the statistics of real and synthesized images due to imperfect rendering. The network achieves state-of-the-art performance on both 2D keypoint estimation and 3D structure recovery. We also show that the recovered 3D information can be used in other vision applications, such as image retrieval.


\end{abstract}

\section{Introduction}
\label{sec:intro}

Deep networks have achieved impressive performance on $1,000$-way image classification~\cite{alexnet}.
However, for any visual system to parse objects in the real world, it needs not only to assign category labels to objects, but also to interpret their intra-class variation.
For example, for a chair, we are interested in its intrinsic properties such as its \emph{style}, \emph{height}, leg \emph{length}, and seat \emph{width}, and extrinsic properties such as its \emph{pose}.

\begin{figure}[t]
	\centering
	\includegraphics[width=\linewidth]{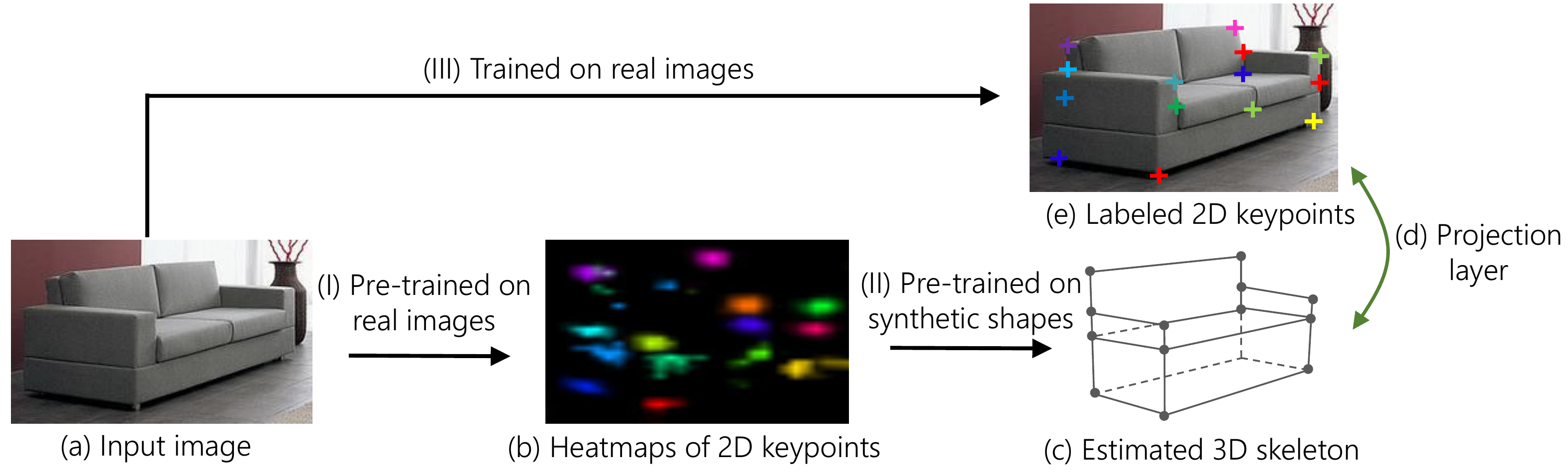}
	\caption{An abstraction of the proposed 3D INterpreter Network (3D-INN). 
	}
	\label{fig:overview}
\end{figure}

In this paper, we recover these object properties from a single image by estimating 3D structure. Instead of a 3D mesh or a depth map~\cite{DB15,Aubry14,prasad2010finding,kar2015category,haosu_sig14,vicente2014reconstructing,huang2015single}, we represent an object via a 3D skeleton~\cite{non_rigid3d}, which consists of keypoints and the connections between them (Figure~\ref{fig:overview}c). 
Being a simple abstraction, the skeleton representation preserves the structural properties that we are interested in.
In this paper, we assume one pre-defined skeleton model for each object category (\eg chair, sofa, and human).

The main challenge of recovering 3D object structure from a single RGB image is the difficulty in obtaining training images with ground truth 3D geometry, as manually annotating 3D structures of objects in real images is labor-intensive and often inaccurate. Previous methods tackle this problem mostly in two ways. 
One is to directly recover a 3D skeleton from estimated 2D keypoint locations by minimizing its reprojection error.
This method uses no training data in 3D reconstruction, thus it is not robust to noisy keypoint estimation, as shown in experiments (\sect{sec:results}). 
The other is to train on synthetically rendered images of 3D objects~\cite{li2015joint,su15}, where complete 3D structure is available.
However, the statistics of synthesized images are often different from those of real images, possibly due to lighting, occlusion, and shape details, making models trained mostly on synthetic data hard to generalize well to real images.

In this paper, we propose \model (\modelshort), an end-to-end framework for recovering 3D object skeletons,
trained on
both real 2D-labeled images and synthetic 3D objects.
Our model has two major innovations.
First, we introduce a \emph{Projection Layer}, a simple renderer which calculates 2D keypoint projections from a 3D skeleton at the end of the network (\fig{fig:overview}d).
This enables \modelshort to predict 3D structural parameters that minimizes the error in the 2D space with labeled real images, without requiring 3D object annotations.

Second,
we further observe that training with real images only under a projection layer is not enough due to
the fundamental ambiguity in 2D-to-3D mapping.
In other words, the algorithm might recover an unnatural 3D geometry whose projection matches the 2D image, because the projection layer only requires the 3D prediction to be plausible in 2D.
We therefore incorporate synthetic 3D objects into training data, in order to encode the knowledge of ``plausible shapes''.
To this end, our model is designed to first predict keypoint locations (\fig{fig:overview}-I) and then to regress 3D parameters (\fig{fig:overview}-II). 
We pre-train the former part with 2D-annotated real images and the latter part with synthetic 3D data, and then train the joint framework end-to-end with the projection layer (\fig{fig:overview}-III). 
We choose heatmaps of keypoints (\fig{fig:overview}b) as an intermediate representation between two components to resolve the domain adaptation issue between real and synthetic data.

Several experiments demonstrate the effectiveness of \modelshort. First, the proposed network achieves state-of-the-art performance on various keypoint localization datasets (FLIC~\cite{FLIC} for human bodies, CUB-200-2011~\cite{CUB} for birds, and our new dataset, \datasetName, for furniture). 
We then evaluate our network on IKEA~\cite{ikea}, a dataset with ground truth 3D object structures and viewpoints. 
On 3D structure estimation, \modelshort shows its advantage over a optimization-based method~\cite{zhou153d} when keypoint estimation is imperfect. 
On 3D viewpoint estimation, it also performs better than the state-of-the-art~\cite{su15}. 
We further evaluate \modelshort, in combination with detection frameworks~\cite{RCNN}, on the popular benchmark PASCAL 3D+~\cite{pascal3d}. Though our focus is not on pose estimation, \modelshort achieves results comparable to the state-of-the-art~\cite{su15,tulsiani2015viewpoints}.
At last, we show qualitatively that \modelshort has wide vision applications including 3D object retrieval.

Our contributions include (1) introducing an end-to-end \model (\modelshort) with a projection layer, which can be trained to predict 3D structural parameters using only 2D-annotated images, (2) using keypoint heatmaps to connect real and synthetic worlds, strengthening the generalization ability of the network, and (3) state-of-the-art performance in 2D keypoint and 3D structure and viewpoint estimation.

\section{Related work}

\xpar{Single image 3D reconstruction} Previous 3D reconstruction methods mainly used object representations based on depth or meshes, or based on skeletons or pictorial structure. Depth-/mesh-based models can recover detailed 3D object structure from a single image, either by adapting existing 3D models from a database~\cite{Aubry14,satkin_bmvc2012,haosu_sig14,huang2015single,zeng20163dmatch,hu2015learning,bansal2016marr,shrivastava2013building,choy20163d}, or by inferring from its detected 2D silhouette~\cite{kar2015category,vicente2014reconstructing,prasad2010finding}. 

In this paper, we choose to use a skeleton-based representation, exploiting the power of abstraction. The skeleton model can capture geometric changes of articulated objects~\cite{non_rigid3d,yasin2016dualsource,akhter2015pose}, like a human body or the base of a swivel chair. Typically, researchers recovered a 3D skeleton from a single image by minimizing its projection error on the 2D image plane~\cite{lowe1987three,leclerc1992optimization,synthesis3d,xue2012example,ramakrishna2012reconstructing,zia2013detailed}. 
Recent work in this line \cite{akhter2015pose,zhou153d} 
demonstrated state-of-the-art performance. In contrast to them, we propose to use neural networks to predict a 3D object skeleton from its 2D keypoints, which is more robust to imperfect detection results and can be jointly learned with keypoint estimators.

Our work also connects to the traditional field of vision as inverse graphics~\cite{hinton1997generative,kulkarni2015deep} and analysis by synthesis~\cite{yuille2006vision,kulkarni2015picture,bever2010analysis,wu2015galileo}, as we use neural nets to decode latent 3D structure from images, and use a projection layer for rendering. Their approaches often required supervision for the inferred representations or made over-simplified assumptions of background and occlusion in images. Our \modelshort learns 3D representation without using 3D supervision, and generalizes to real images well.

\vpar{2D keypoint estimation} Another line of related work is 2D keypoint estimation. During the past decade, researchers have made significant progress in estimating keypoints on humans~\cite{FLIC,yang2011articulated} and other objects~\cite{CUB,shih2015part}. Recently, there have been several attempts to apply convolutional neural networks to human keypoint estimation~\cite{toshev2014deeppose,tompson2015efficient,carreira2015human,newell2016stacked}, which all achieved significant improvement. Inspired by these work, we use 2D keypoints as our intermediate representation, and aim to recover 3D skeleton from them.

\vpar{3D viewpoint estimation} 
3D viewpoint estimation seeks to estimate the 3D orientation of an object from a single image~\cite{pascal3d}. Some previous methods formulated it as a classification or regression problem, and aimed to directly estimate the viewpoint from an image~\cite{urtasun_3d_car,su15}. Others proposed to estimate 3D viewpoint from detected 2D keypoints or edges in the image~\cite{zia2013detailed,fpm,tulsiani2015viewpoints}. While the main focus of our work is to estimate 3D object structure, our method can also predict its 3D viewpoint. 

\vpar{Training with synthetic data} 
Synthetic data are often used to augment the training 
set~\cite{haosu_sig14,peng2014exploring,shakhnarovich2003fast}. 
Su \etal~\cite{haosu_sig14} attempted to train a 3D viewpoint estimator using a combination of real and synthetic images, while Sun \etal~\cite{sun2014virtual} and Zhou \etal~\cite{zhou2016learning} also used a similar strategy for object detection and matching, respectively. Huang \etal~\cite{huang2015single} 
analyzed the invariance of convolutional neural networks using synthetic images. For image synthesis, Dosovitskiy \etal~\cite{DB15} trained a neural network to generate new images using synthetic images.

In this paper, we combine real 2D-annotated images and synthetic 3D data for training \modelshort to recover a 3D skeleton. We use heatmaps of 2D keypoints, instead of (often imperfectly) rendered images, from synthetic 3D data, so that our algorithm has better generalization ability as the effects of imperfect rendering are minimized. Yasin \etal~\cite{yasin2016dualsource} also proposed to use both 2D and 3D data for training, but they uses keypoint location, instead of heatmaps, as the intermediate representation that connects 2D and 3D.

\section{Methods} \label{sec:methods}

\begin{figure}[t]
	\centering
	\includegraphics[width=\linewidth]{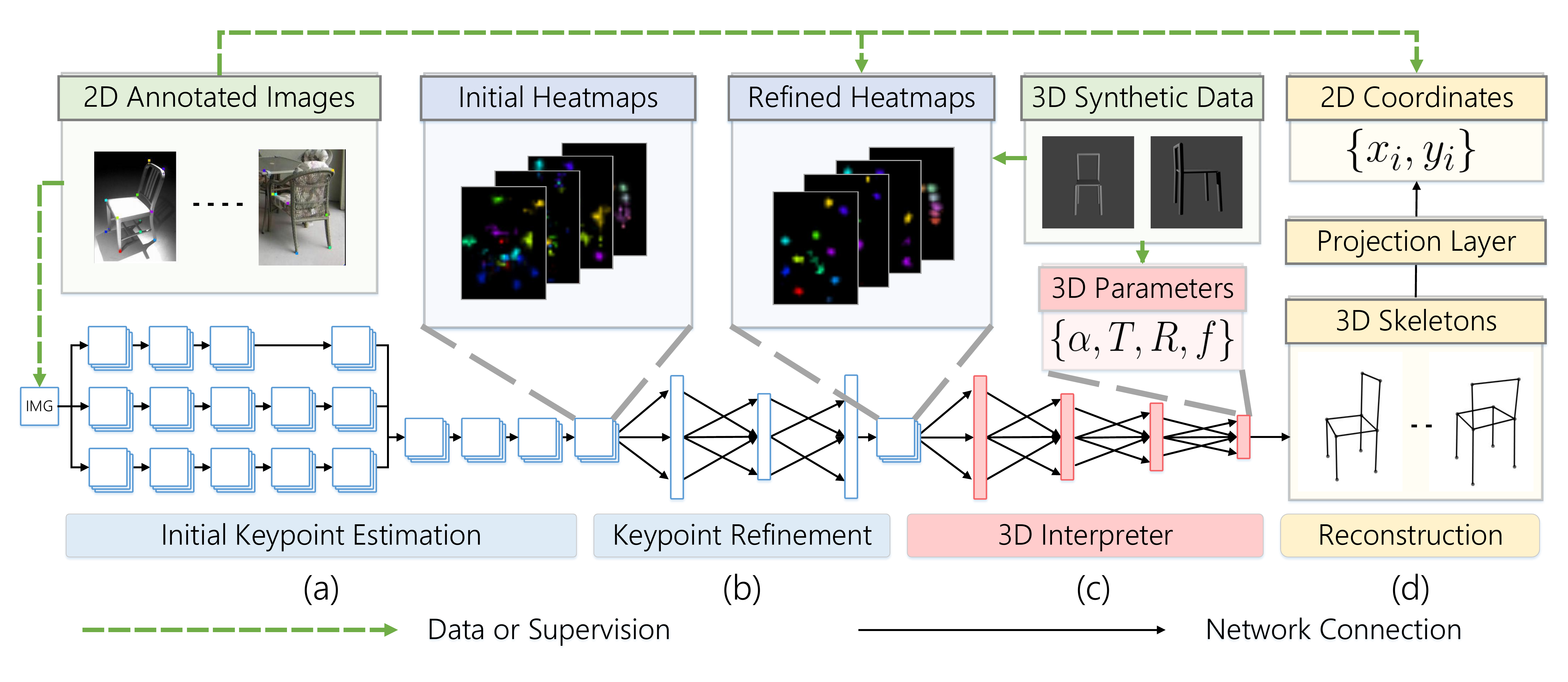}
	\caption{\modelshort takes a single image as input and reconstructs the detailed 3D structure of the object in the image (\eg, human, chair, \etc). The network is trained independently for each category, and here we use chairs as an example. \textbf{(a)} Estimating 2D keypoint heatmaps with a multi-scale CNN. \textbf{(b)} Refining keypoint locations by considering the structural constraints between keypoints. This is implicitly enforced with an information bottleneck which yields cleaner heatmaps. \textbf{(c)} Recovered 3D structural and camera parameters $\{\alpha, T, R, f\}$. \textbf{(d)} The projection layer maps reconstructed 3D skeletons back to 2D keypoint coordinates.} 
	\label{fig:architecture} 
\end{figure}

We design a deep convolutional network to recover 3D object structure. 
The input to the network is a single image with an object of interest at its center, which can be obtained by state-of-the-art object detectors~\cite{FastRCNN}. 
The output of the network is a 3D object skeleton, 
including its 2D keypoint locations, 3D structural parameters, and 3D poses (see Figure \ref{fig:architecture}). 
In the following, we will describe our 3D skeleton representation (Section~\ref{sec:skeleton}), network architecture (Section~\ref{sec:network}), and training strategy (Section~\ref{sec:synthetic_data}).

\begin{figure}[t]
	\centering
    \includegraphics[width=0.6\columnwidth]{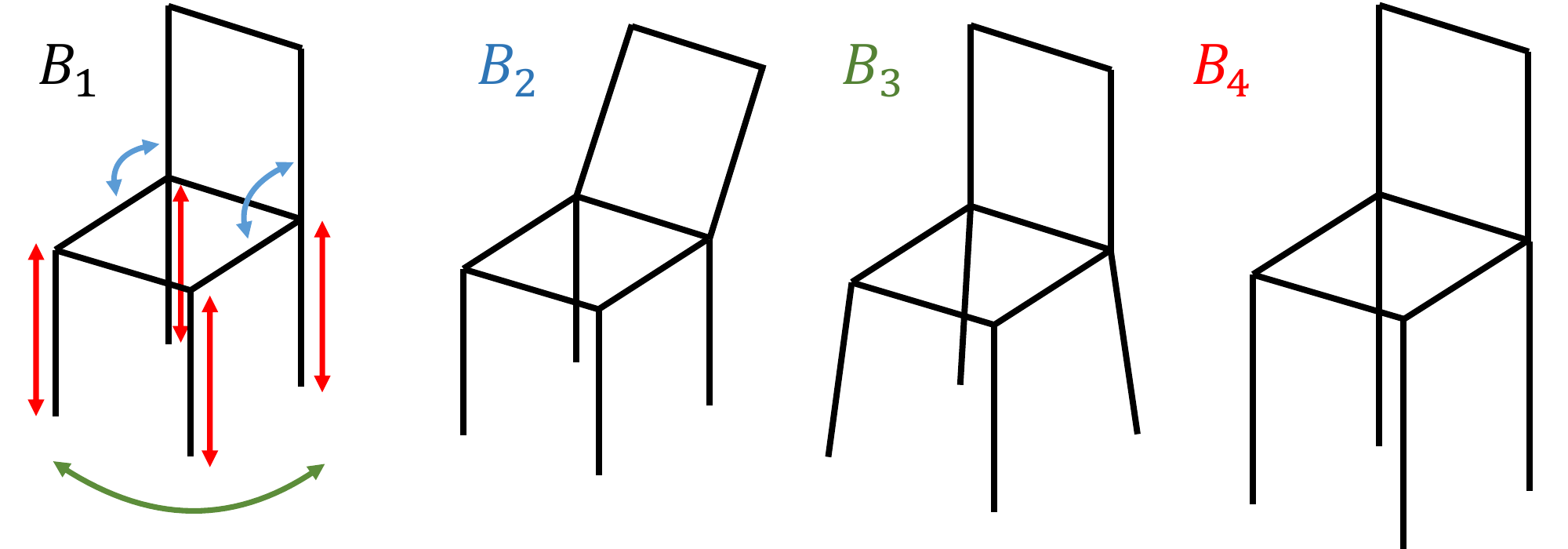}
	\caption{A simplification of our skeleton model and base shapes for chairs}
	\label{fig:syn}
\end{figure}

\subsection{3D Skeleton Representation \label{sec:skeleton}}

As discussed in \sect{sec:intro}, we use skeletons as our 3D object representation. A skeleton consists of a set of keypoints as well as their connections. For each object category, we manually design a 3D skeleton characterizing its abstract 3D geometry. 

There exist intrinsic ambiguities in recovering 3D keypoint locations from a single 2D image.
We resolve this issue by assuming that objects can only have constrained deformations~\cite{non_rigid3d}.
For example, chairs may have various leg lengths, but for a single chair, its four legs are typically of equal length.
We model these constraints by formulating 3D keypoint locations as a weighted sum of a set of base shapes~\cite{kar2015category}. The first base shape is the mean shape of all objects within the category, and the rest define possible deformations and intra-class variations. \fig{fig:syn} shows an simplification of our skeleton representation for chairs: the first is the mean shape of chairs, the second controls how the back bends, and the last two are for legs.
The weight for each base shape determines how strong the deformation is, and we denote these weights as the \emph{internal parameters} of an object.

Formally, let $\cY \in \bR^{3\times N}$ be a matrix of 3D coordinates of all $N$ keypoints. Our assumption is that the 3D keypoint locations are a weighted sum of base shapes $B_k\in\bR^{3\times N}$, or $\cY = \sum_{k=1}^K \alpha_k B_k$,
where $\{\alpha_k\}$ is the set of internal parameters of this object and $K$ is the number of base shapes. 

Further, let $\cX \in \bR^{2\times N}$ be the corresponding 2D coordinates. Then the relationship between the observed 2D coordinates $\cX$ and the internal parameters $\{\alpha_k\}$ is
\begin{equation}
\cX = P(R\cY + T) = P(R\sum_{k=1}^K \alpha_k B_k + T),
\label{eq:pnp}
\end{equation}
where $R\in\bR^{3\times 3}$ (rotation) and $T\in\bR^{3}$ (translation) are the external parameters of the camera, and $P$ is a projective transformation. $P$ only depends on the focal length $f$ under the central projection we assuming.

Therefore, to recover the 3D structural information of an object in a 2D image, we only need to estimate its internal parameters ($\{\alpha_k\}$) and the external viewpoint parameters ($R$, $T$, and $f$). In the following section, we discuss how we design a neural network for this task, and how it can be jointly trained with real 2D images and synthetic 3D objects.

\subsection{Architecture of \modelshort\label{sec:network}}
Our network consists of three components: first, a keypoint estimator, which localizes 2D keypoints of objects from 2D images by regressing to their heatmaps (\fig{fig:architecture}a and b, blue part); second, a 3D interpreter,
which infers internal 3D structural and viewpoint parameters from the heatmaps (\fig{fig:architecture}c, red part);
third, a projection layer, mapping 3D skeletons to 2D keypoint locations so that real 2D-annotated images can be used as supervision (\fig{fig:architecture}d, yellow part).

\vpar{Keypoint Estimation}
The keypoint estimation component consists of two steps: initial estimation (\fig{fig:architecture}a) and keypoint refinement (\fig{fig:architecture}b). 

The network architecture for initial keypoint estimation is inspired by the pipeline proposed by Tompson \etal~\cite{tompson2014joint,tompson2015efficient}. The network takes multi-scaled images as input and estimates keypoint heatmaps. Specifically, we apply Local Contrast Normalization (LCN) on each image, and then scale it to $320\times240$, $160\times120$, and $80\times60$ as input to three separate scales of the network. The output is $k$ heatmaps, each with resolution $40\times30$, where $k$ is the number of keypoints of the object in the image. 

At each scale, the network has three sets of $5\times5$ convolutional (with zero padding), ReLU, and $2\times2$ pooling layers, followed by a $9\times9$ convolutional and ReLU layer. The final outputs for the three scales are therefore images with resolution $40\times30$, $20\times15$, and $10\times7$, respectively. We then upsample the outputs of the last two scales to ensure they have the same resolution ($40\times30$). The outputs from the three scales are later summed up and sent to a Batch Normalization layer and three $1\times1$ convolution layers, whose goal is to regress to target heatmaps. We found that Batch Normalization is critical for convergence, while Spatial Dropout, proposed in~\cite{tompson2015efficient}, does not affect performance.

The second step of keypoint estimation is keypoint refinement, whose goal is to implicitly learn category-level structural constraints on keypoint locations after the initial keypoint localization. The motivation is to exploit the contextual and structural knowledge among keypoints (\eg, arms cannot be too far from the torso). We design a mini-network which, 
like an auto-encoder, has information bottleneck layers, enforcing it to implicitly model the relationship among keypoints. Some previous works also use this idea and achieve better performance with lower computational cost in object detection~\cite{FastRCNN} and face recognition~\cite{deepface2}.

In the keypoint refinement network, We use three fully connected layers with widths $8,192$, $4,096$, and $8,192$, respectively. After refinement, the heatmaps of keypoints are much cleaner, as shown in \fig{fig:sl} and \sect{sec:results}. 

\vpar{3D Interpreter}
The goal of our 3D interpreter is to infer 3D structure and viewpoint parameters, using estimated 2D heatmaps from earlier layers. While there are many different ways of solving Equation~\ref{eq:pnp}, our deep learning approach has clear advantages. First, traditional methods~\cite{hejrati2012analyzing,non_rigid3d} 
that minimize the reprojection error consider only one keypoint hypothesis, and is therefore not robust to noises in keypoint detection. 
In contrast, our framework uses soft heatmaps of keypoint locations, as shown in \fig{fig:architecture}c, which is more robust when some keypoints are invisible or incorrectly located. 
Further, our algorithm only requires a single forward propagation during testing, making it more efficient than the most previous optimization-base methods.

As discussed in \sect{sec:skeleton}, the set of 3D parameters we estimate consists of $S=\{\alpha_i, R, T, f\}$, with which we are able to recover the 3D object structure using Equation~\ref{eq:pnp}. As shown in Figure~\ref{fig:architecture}c, we use four fully connected layers as our 3D interpreter, with widths $2,048$, $512$, $128$, and $|S|$, respectively. 
The Spatial Transformer Network~\cite{jaderberg2015spatial} also explored the idea of learning rotation parameters $R$ with neural nets, but our network can also recover structural parameters $\{\alpha_i\}$. Note that our representation for latent parameters may also be naturally extended to other types of abstract 3D representations.

\vpar{Projection Layer}
The last component of the network is a projection layer (\fig{fig:architecture}d). The projection layer takes estimated 3D parameters as input, and computes projected 2D keypoint coordinates $\{x_i,y_i\}$ using Equation~\ref{eq:pnp}. As all operations are differentiable, the projection layer enables us to use 2D-annotated images as ground truth, and run back-propagation to update the entire network.

\subsection{Training Strategy}\label{sec:synthetic_data}

A straightforward training strategy is to use real 2D images as input, and their 2D keypoint locations as supervision for the output of the projection layer. Unfortunately, experiments show that the network can hardly converge using this training scheme, due to the high-dimensional search space and the ambiguity in the 3D to 2D projection. 

We therefore adopt an alternative three-step training strategy: first, training the keypoint estimator (\fig{fig:architecture}a and \ref{fig:architecture}b) using real images with 2D keypoint heatmaps as supervision; second, training the 3D interpreter (\fig{fig:architecture}c) using synthetic 3D data as there are no ground truth 3D annotations available for real images; and third, training the whole network using real 2D images with supervision on the output of the projection layer at the end.

To generate synthetic 3D objects, for each object category, we first randomly sample structural parameters $\{\alpha_i\}$ and viewpoint parameters $P$, $R$ and $T$. Then we calculate 3D keypoint coordinates using Equation~\ref{eq:pnp}. To model deformations that cannot be captured by base shapes, we add Gaussian perturbation to 3D keypoint locations of each synthetic 3D object, whose variance is $1\%$ of its diagonal length. 
Examples of synthetic 3D shapes are shown in~\fig{fig:architecture}c. Note that we are not rendering these synthesized objects, as we are only using heatmaps of keypoints, rather than rendered images, as training input.

\section{Evaluation} \label{sec:results}

We evaluate our entire framework, \modelshort, as well as each component within. In this section, we 
present both qualitative and quantitative results on 2D keypoint estimation (\sect{sec:eval_2d}) and 3D structure recovery (\sect{sec:eval_3d}).

\begin{figure}[t!]
	\centering
	\includegraphics[width=\linewidth]{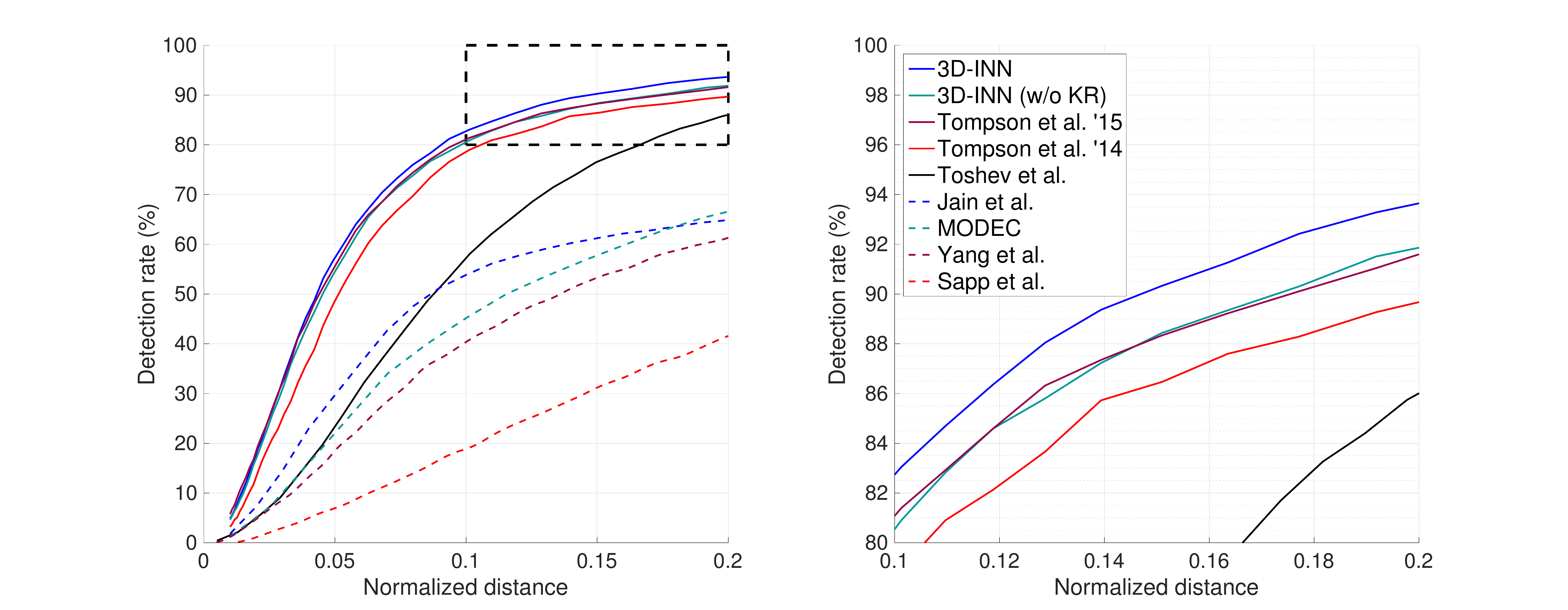}
	\caption{\label{fig:human_keypoint_curves} PCK curves on the FLIC dataset. \modelshort performs consistently better than other methods. Without keypoint refinement, it is comparable to Tompson \etal~\cite{tompson2015efficient}. A zoomed view of the dashed rectangle is shown on the right.}
\end{figure}

\subsection{2D Keypoint Estimation}
\label{sec:eval_2d}

\xpar{Data}
For 2D keypoint estimation, we evaluate our algorithm on three image datasets: FLIC~\cite{FLIC} for human bodies, CUB-200-2011~\cite{CUB} for birds, and a new dataset \datasetName for furniture. Specifically, FLIC is a challenging dataset containing $3,987$ training images and $1,016$ test images, each labeled with $10$ keypoints of human bodies. The CUB-200-2011 dataset was originally proposed for fine-grained bird classification, but with labeled keypoints of bird parts. It has $5,994$ images for training and $5,794$ images for testing, each coming with up to $15$ keypoints.

We also introduce a new dataset, \datasetName, which contains five categories: bed, chair, sofa, swivel chair, and table. There are $1,000$ to $2,000$ images in each category, where $80\%$ are for training and $20\%$ for testing. For each image, we asked three workers on Amazon Mechanical Turk to label locations of a pre-defined category-specific set of keypoints; we then, for each keypoint, used the median of the three responses as ground truth.

\vpar{Metrics}
To quantitatively evaluate the accuracy of estimated keypoints on FLIC (human body), we use the standard Percentage of Correct Keypoints (PCK) measure~\cite{FLIC} to be consistent with previous works~\cite{FLIC,tompson2014joint,tompson2015efficient}. We use the evaluation toolkit and results of competing methods released by the Tompson \etal~\cite{tompson2015efficient}. On CUB-200-2011 (bird) and the new \datasetName (furniture) dataset, following the convention~\cite{liu2013bird,shih2015part}, we evaluate results in Percentage of Correct Parts (PCP) and Average Error (AE). PCP is defined as the percentage of keypoints localized within $1.5$ times of the standard deviation of annotations. We use the evaluation code from~\cite{liu2013bird} to ensure consistency. Average error is computed as the mean of the distance, bounded by $5$, between a predicted keypoint location and ground truth. 

\begin{figure}[t!]
    \centering
	\includegraphics[width=0.85\linewidth]{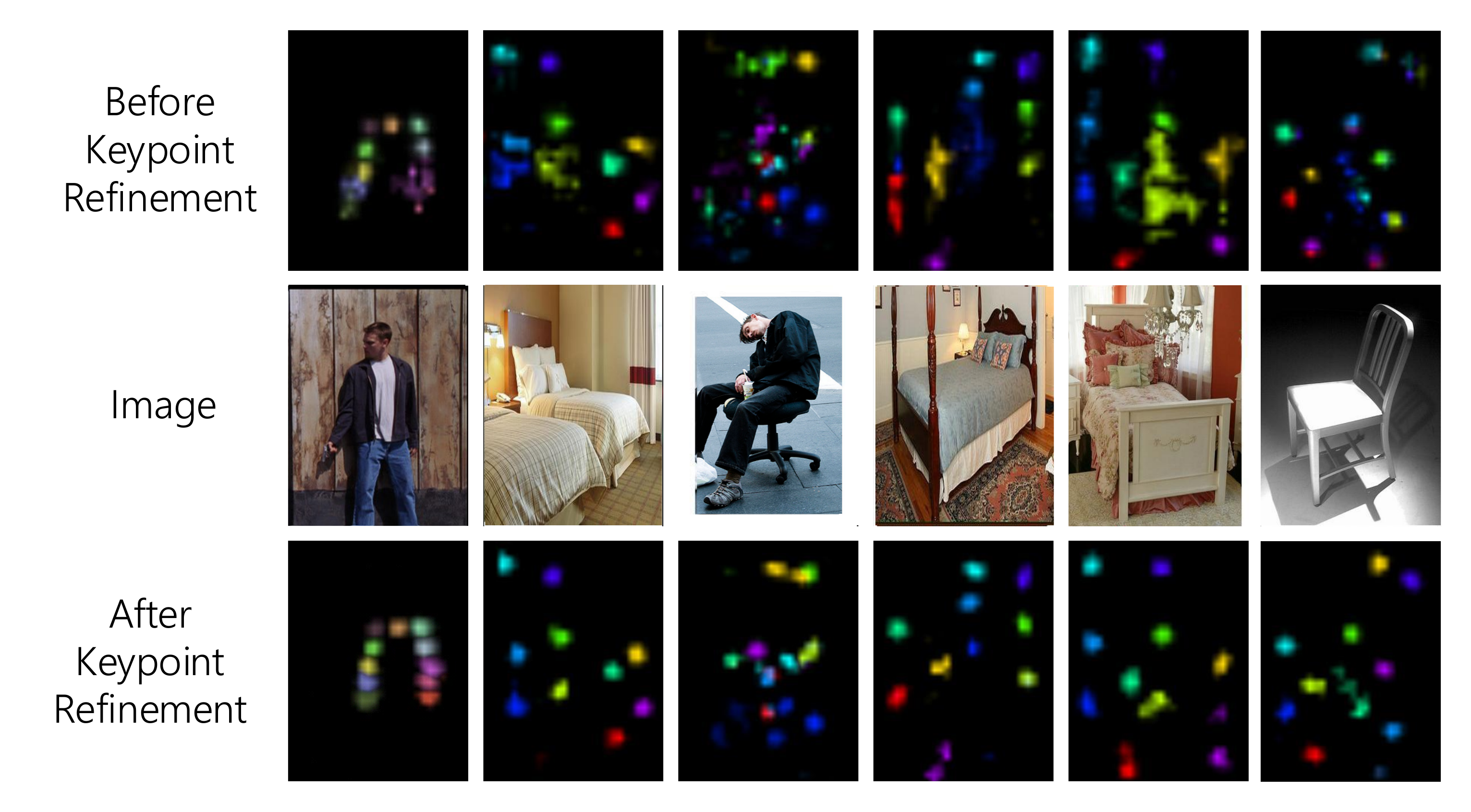}
    \caption{2D keypoint predictions from a single image, where each color corresponds to a keypoint. The keypoint refinement step cleans up false positives and produces more regulated predictions. }
    \label{fig:sl}
\end{figure}

\begin{table}[t!]
    \centering
 
    \begin{tabular}{lcccccc}
    \toprule
    Method & Poselets~\cite{poselet} & Consensus~\cite{consensus} & Exemplar~\cite{liu2013bird} & Mdshift~\cite{shih2015part} & \modelshort & Human\\
    \midrule
    PCP (\%) & 27.47 & 48.70 & 59.74 & {\bf 69.1} & 66.7 & 84.72 \\
    Average Error & 2.87 & 2.13 & 1.80 & 1.39 & {\bf 1.35} & 1.00 \\
    \bottomrule
    \end{tabular}
    \vspace{5pt}
    \caption{Keypoint estimation results on CUB-200-2011, measured in PCP (\%) and AE. Our method is comparable to Mdshift~\cite{shih2015part} (better in AE but worse in PCP), and better than all other algorithms.}
    \label{table:bird_keypoint_result}
\end{table}

\begin{table}[t!]
    \centering
    \begin{tabular}{L{1.2cm}lcccc}
    \toprule
     & Method & Bed & Chair & Sofa & Swivel Chair \\
    \midrule
    \multirow{2}{*}{PCP} & \modelshort & \tb{77.4} & \tb{87.7} & \tb{77.4} & \tb{78.5} \\
    & Tompson \etal~\cite{tompson2015efficient} & 76.2 & 85.3 & 76.9 & 69.2 \\
    \midrule
    \multirow{2}{*}{AE} & \modelshort & \tb{1.16} & \tb{0.92} & \tb{1.14} & \tb{1.19} \\
    & Tompson \etal~\cite{tompson2015efficient} & 1.20 & 1.02 & 1.19 & 1.54 \\
    \bottomrule
    \end{tabular}%}
    \vspace{5pt}
    \caption{Keypoint estimation results of \modelshort and Tompson~\etal~\cite{tompson2015efficient} on \datasetName, measured in PCP (\%) and AE.  \modelshort is consistently better in both measures. We retrained the network in ~\cite{tompson2015efficient} on \datasetName.}
    \label{table:other_keypoint}
\end{table}

\vpar{Results}
For 2D keypoint detection, we only train the keypoint estimator in our \modelshort (\fig{fig:architecture}a and \ref{fig:architecture}b) using the training images in each dataset. \fig{fig:human_keypoint_curves} shows the accuracy of keypoint estimation on the FLIC dataset. On this dataset, we employ a fine-level network for post-processing, as suggested by~\cite{tompson2015efficient}. Our method performs better than all previous methods~\cite{FLIC,tompson2014joint,tompson2015efficient,yang2011articulated,toshev2014deeppose} at all precisions. Moreover, the keypoint refinement step improves results significantly (about $2\%$ for a normalized distance $\geq0.15$), without which our framework has similar performance with~\cite{tompson2015efficient}. Such improvement is also demonstrated in Figure~\ref{fig:sl}, where the heatmaps after refinement are far less noisy.

The accuracy of keypoint estimation on CUB-200-201 dataset is listed in Table~\ref{table:bird_keypoint_result}. Our method is better than~\cite{liu2013bird} in both metrics, and is comparable to the state-of-the-art~\cite{shih2015part}. Specifically, compared with~\cite{shih2015part}, our model more precisely estimates the keypoint locations for correctly detected parts (a lower AE), but miss more parts in the detection (a lower PCP). On our \datasetName dataset, our model achieves higher PCPs and lower AEs compared to the state-of-the-art~\cite{tompson2015efficient} for all categories, as shown in Table~\ref{table:other_keypoint}. These experiments in general demonstrate the effectiveness of our model on keypoint detection.

\subsection{Structural Parameter Estimation}
\label{sec:eval_3d}

For 3D structural parameter estimation, we evaluate \modelshort from three different prospectives. First, we evaluate our 3D interpreter (\fig{fig:architecture}c alone) against the optimization-based method~\cite{zhou153d}. Second, we test our full pipeline on the IKEA dataset~\cite{ikea}, where ground truth 3D labels are available. We show qualitative results on three datasets: \datasetName, IKEA, and SUN~\cite{SUN} at last.

\vpar{Comparing with an optimization-based method}
We first compared our 3D interpreter (\fig{fig:architecture}c) with the  state-of-the-art optimization-based method that directly minimizing re-projection error (Equation~\ref{eq:pnp}) on the synthetic data. 

We first tested the effectiveness of our 3D interpreter (\fig{fig:architecture}c) on synthetic data. We compare our trained 3D interpreter against the state-of-the-art method on directly minimizing re-projection error (Equation~\ref{eq:pnp}). Since most optimization based methods only consider the parallel projection, we extend the one by Zhou~\etal~\cite{zhou153d} as follows. We first uses their algorithm to get an initial guess of internal parameters and viewpoints, and then applying a simple gradient descent method to refine it considering perspective distortion.

We generate synthetic data for this experiment, using the scheme described in~\sect{sec:synthetic_data}. Each data point contains the 2D keypoint heatmaps of an object, and its corresponding 3D keypoint locations and viewpoint, which we would like to estimate. We also add different levels of salt-and-pepper noise to heatmaps to evaluate the robustness of both methods. We generated 30,000 training and 1,000 testing cases. Because the analytical solution only takes keypoint coordinates as input, we convert heatmaps to coordinates using argmax.

\begin{figure}[t]
    \centering
            \includegraphics[width=\linewidth]{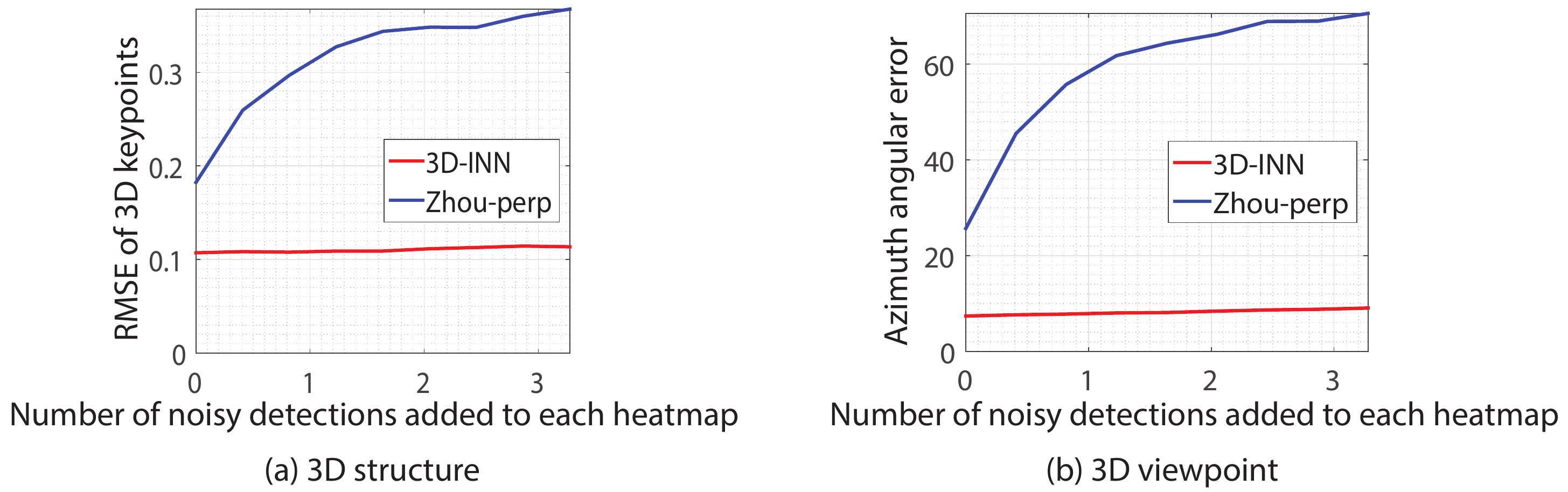}\\
    \caption{Plots comparing our method against an analytic solution on synthetic heatmap. 
    (a) The accuracy of 3D structure estimation; (b) The accuracy of 3D viewpoint estimation.
    }
    \label{fig:plot_syn}
\end{figure}

For both methods, we evaluate their performance on both 3D structure recovery and 3D viewpoint estimation. For 3D structure estimation, we compare their accuracies on 3D keypoint estimation ($\cY$ in Section~\ref{sec:skeleton}); for 3D viewpoint estimation, we evaluate errors in azimuth angle, following previous work~\cite{su15}. As the original algorithm by Zhou~\etal~\cite{zhou153d} was mainly designed for the parallel projection and comparatively clean heatmaps, our 3D interpreter outperforms it in the presence of noise and perspective distortion, as shown in~\fig{fig:plot_syn}.

\vpar{Evaluating the full pipeline}
We now evaluate \modelshort on estimating 3D structure and 3D viewpoint. We use the IKEA dataset~\cite{ikea} for evaluation, as it provides ground truth 3D mesh models and the associated viewpoints for testing images. We manually label ground truth 3D keypoint locations on provided 3D meshes, and calculate the root-mean-square error (RMSE) between estimated and ground truth 3D keypoint locations. 

As IKEA only have no more than $200$ images per category, we instead train \modelshort on our \datasetName, as well as one million synthetic data points, using the strategy described in ~\sect{sec:synthetic_data}. Note that, first, we are only using no more than $2,000$ real images per category for training and, second, we are testing the trained model on different datasets, avoiding the possible dataset bias~\cite{datasetbias}.

\begin{figure}[t]
	\centering
    \begin{tabular}{c}
    \includegraphics[width=\columnwidth]{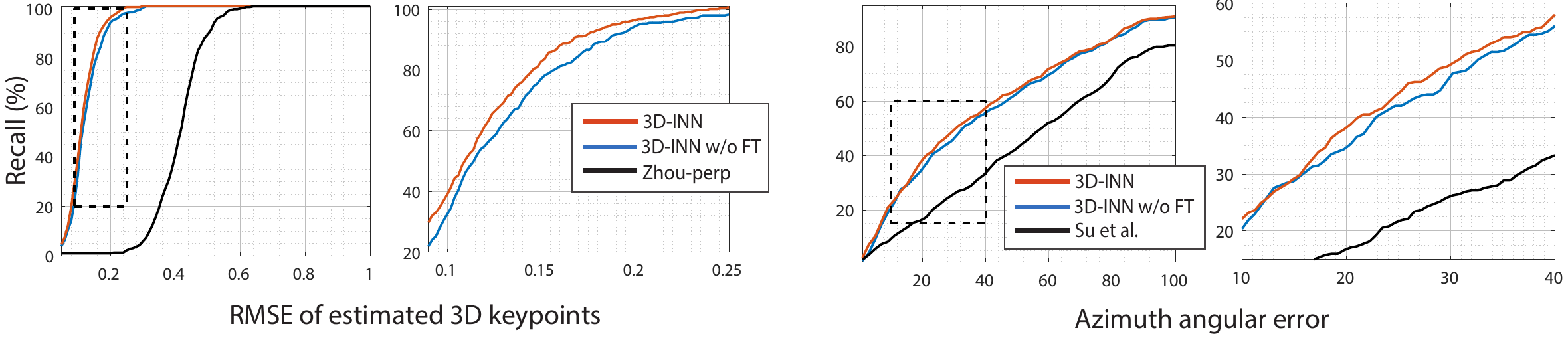} \\
    \begin{tabular}{cC{0.3cm}c}
    Average recall \% && Average recall \% \\
    \scalebox{0.9}{
\begin{tabular}{lcccc} \toprule
Method& Bed& Sofa& Chair& Avg.\\ \midrule 
3D-INN & \textbf{88.64} & \textbf{88.03} & \textbf{87.84} & \textbf{88.03} \\ 
3D-INN w/o FT & 87.19 & 87.10 & 87.08 & 87.10 \\ 
Zhou-perp~\cite{zhou153d} & 52.31 & 58.02 & 60.76 & 58.46 \\ 
\bottomrule \end{tabular}
    } &&
    \scalebox{0.9}{
\begin{tabular}{lcccc} \toprule
Method& Table& Sofa& Chair& Avg.\\ \midrule 
3D-INN & \textbf{55.02} & 64.65 & \textbf{63.46} & \textbf{60.30} \\ 
3D-INN w/o FT & 52.33 & \textbf{65.45} & 62.01 & 58.90 \\ 
Su~\etal~\cite{su15} & 52.73 & 35.65 & 37.69 & 43.34 \\ 
\bottomrule \end{tabular}
    } \\ \\
    (a) Structure estimation && (b) Pose estimation\\
    \end{tabular}
    \end{tabular}
	\caption{\label{fig:ikea} Evaluation on the IKEA dataset~\cite{ikea}. (a) The accuracy of structure estimation. RMSE-Recall curved is shown in the first row, and zoomed-views of the dashed rectangular regions are shown on the right. The third row shows the average recall on all thresholds. (b) The accuracy of pose estimation. \
	}
\end{figure}

The left half of \fig{fig:ikea} shows RMSE-Recall curve of both our algorithm and the optimization-based method described above (Zhou-perp~\cite{zhou153d}). The $y$-axis shows the recall --- the percentage of testing samples under a certain RMSE threshold. We test two versions of our algorithm: with fine-tuning (3D-INN) and without fine-tuning (3D-INN w/o FT). Both significantly outperform the optimization-based method~\cite{zhou153d}, as ~\cite{zhou153d} is not designed for multiple keypoint hypothesis and perspective distortion, while our \modelshort can deal with them. Also, finetuning improves the accuracy of keypoint estimation by about $5\%$ under the RMSE threshold $0.15$.

\begin{table}[t!]
\footnotesize
\centering
\begin{tabular}{L{1.5cm}ccccc}
\toprule
Category & \; VDPM~\cite{pascal3d} & \; DPM-VOC+VP~\cite{pepik2012teaching} & \; Su~\etal~\cite{su15} & \; V\&K~\cite{tulsiani2015viewpoints} & \; 3D-INN \\ 
\midrule
Chair & 6.8 & 6.1 & 15.7 & \textbf{25.1} & 23.1 \\
Sofa & 5.1 & 11.8 & 18.6 & 43.8 & \textbf{45.8} \\
\bottomrule
\end{tabular}
\vspace{5pt}
\caption{Joint object detection and viewpoint estimation on PASCAL 3D+~\cite{pascal3d}. Following previous work, we use Average Viewpoint Precision (AVP) as our measure, which extends AP so that a true positive should have both a correct bounding box and a correct viewpoint (here we use a 4-view quantization). Both V\&K~\cite{tulsiani2015viewpoints} and our algorithm (\modelshort) use R-CNN~\cite{RCNN} for object detection, and others use their own detection algorithm. VDPM~\cite{pascal3d} and DPM-VOC+VP~\cite{pepik2012teaching} are trained on PASCAL VOC 2012, V\&K~\cite{tulsiani2015viewpoints} is trained on PASCAL 3D+, Su~\etal~\cite{su15} is trained on PASCAL VOC 2012, together with synthetic 3D CAD models, and \modelshort is trained on \datasetName.}
\label{table:pascal3d}
\end{table}

\begin{figure}[ht!]
	\begin{minipage}{\linewidth}
	{\centering
	\includegraphics[width=0.9\linewidth]{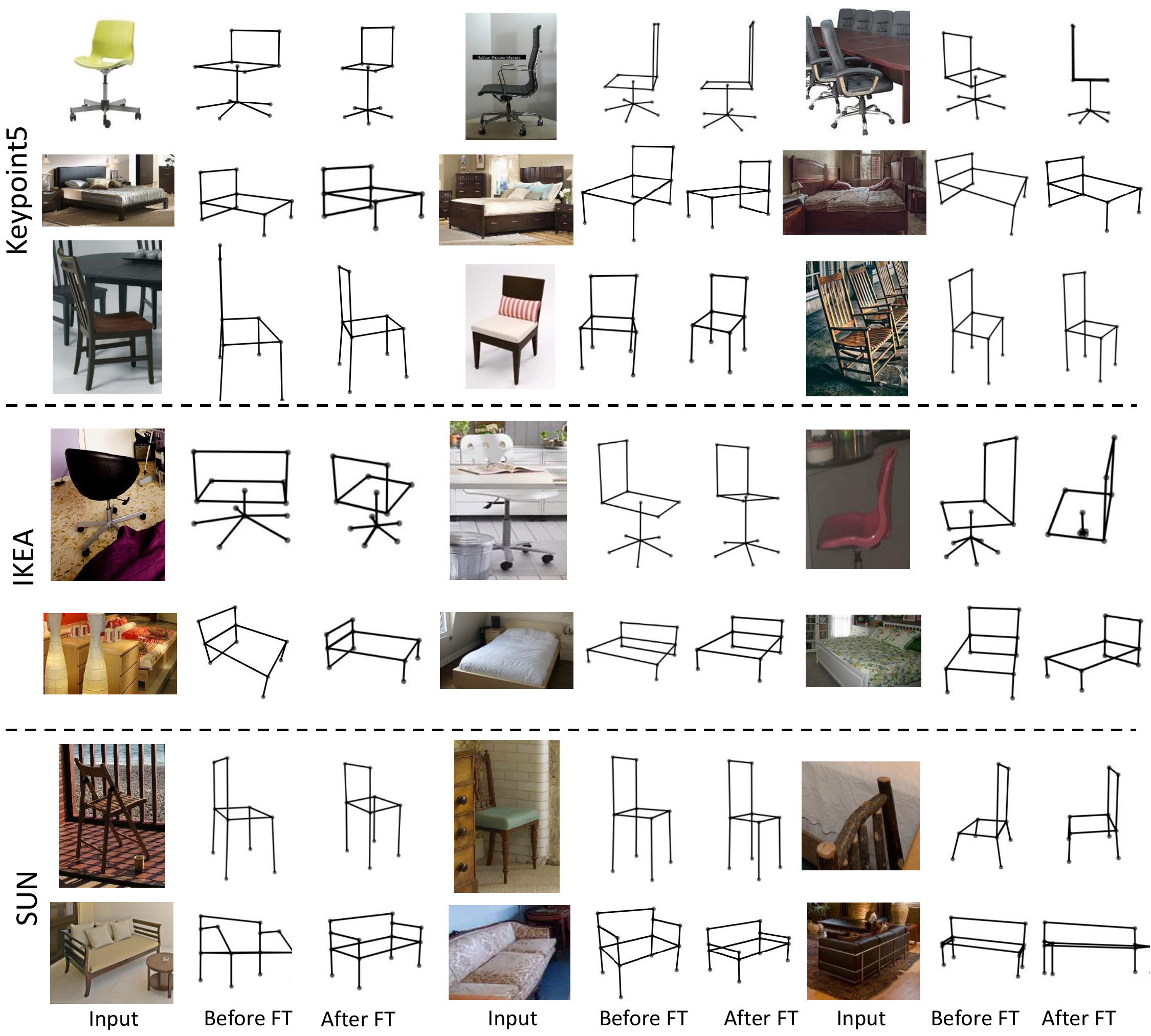}
	\caption{\label{fig:res} Qualitative results on \datasetName, IKEA, and SUN databases. For each example, the first one is the input image, the second one is the reconstruct 3D skeleton using the network before fine-tuning, and third one is using the network after fine-tuning. The last column shows failure cases.}
	}
	\end{minipage}
	\begin{minipage}{\linewidth}
	\vspace{10pt}
\begin{tabular}{ccc}
\includegraphics[width=0.44\linewidth]{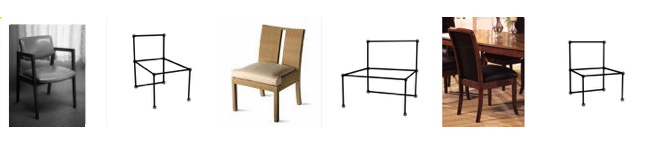} &~~~~&
\includegraphics[width=0.44\linewidth]{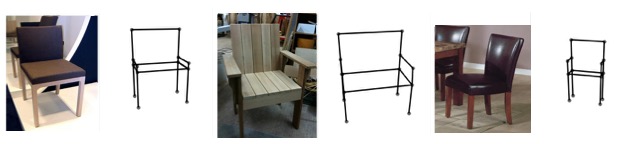} \\
Training: beds, Test: chairs &~~~~~~~~& Training: sofas, Test: chairs
\end{tabular}
\caption{Qualitative results on chairs using networks trained on sofas or beds. In most cases models provide reasonable output. Mistakes are often due to the difference between the training and test sets, \eg, in the third example, the model trained on beds fails to estimate chairs facing backward.}
\label{fig:cross_category}
	\end{minipage}
\end{figure}

Though we focus on recovering 3D object structure, as an extension, we also evaluate \modelshort on 3D viewpoint estimation. We compare it with the state-of-the-art viewpoint estimation algorithm by Su~\etal~\cite{su15}. The right half of \fig{fig:ikea} shows the results (recall) in azimuth angle. As shown in the table, \modelshort outperforms Su~\etal~\cite{su15} by about $40\%$ (relative), measured in average recall. This is mainly because it is not straightforward for Su~\etal~\cite{su15}, mostly trained on (cropped) synthesized images, to deal with the large number of heavily occluded objects in the IKEA dataset.

Although our algorithm assumes a centered object in an input image, we can apply it, in combination with an object detection algorithm, on images where object locations are unknown.
We evaluate the results of joint object detection and viewpoint estimation on PASCAL 3D+ dataset~\cite{pascal3d}. We use the standard R-CNN~\cite{RCNN} for object detection, and our \modelshort for viewpoint estimation. Table~\ref{table:pascal3d} shows that our model is comparable with the state-of-the-art~\cite{tulsiani2015viewpoints}, and ourperforms other algorithms with a significant margin. Note that all the other algorithms are trained on either PASCAL VOC or PASCAL 3D+, while our algorithm is trained on \datasetName, which indicates that our learned model is not suffering much from the dataset bias problem~\cite{datasetbias}.

\vpar{Qualitative results on benchmarks}
At last, we show qualitative results on \datasetName, IKEA, and the SUN database~\cite{SUN} in Figure~\ref{fig:res}. When the image is clean and objects are not occluded, our algorithm can recover 3D object structure and viewpoint with high accuracy, while fine-tuning can further helps to improve the results (see chairs at row 1 column 1, and row 4 column 1).
Our algorithm is also robust of partial occlusion, demonstrated by the IKEA bed at row 5 column 1. One major failure case is when the object is heavily cropped in the input image (see the last column in row 4 to 7), as the 3D object skeleton becomes hard to infer. 

When \modelshort is used in combination with detection models, it needs to deal with imperfect detection results. Here, we also evaluate \modelshort on noisy input, specifically, on images with an object from a different but similar category. Figure~\ref{fig:cross_category} shows the recovered 3D structures of chairs using a model trained either on sofas or beds. In most cases \modelshort still provides reasonable output, and the mistakes are mostly due to the difference between training and test sets, \eg, the model trained on beds does not perform well on chairs facing backward, because there are almost no beds with a similar viewpoint in the training set.

\section{Applications}

Our inferred latent parameters, as a compact and informative representation of objects in images, have wide applications. In this section, we demonstrate representative ones including image retrieval and object graph construction.

\begin{figure}[t]
    \centering
    \begin{minipage}{0.4\linewidth}
    \includegraphics[trim={22cm 1.2cm 0 0},clip,width=\linewidth]{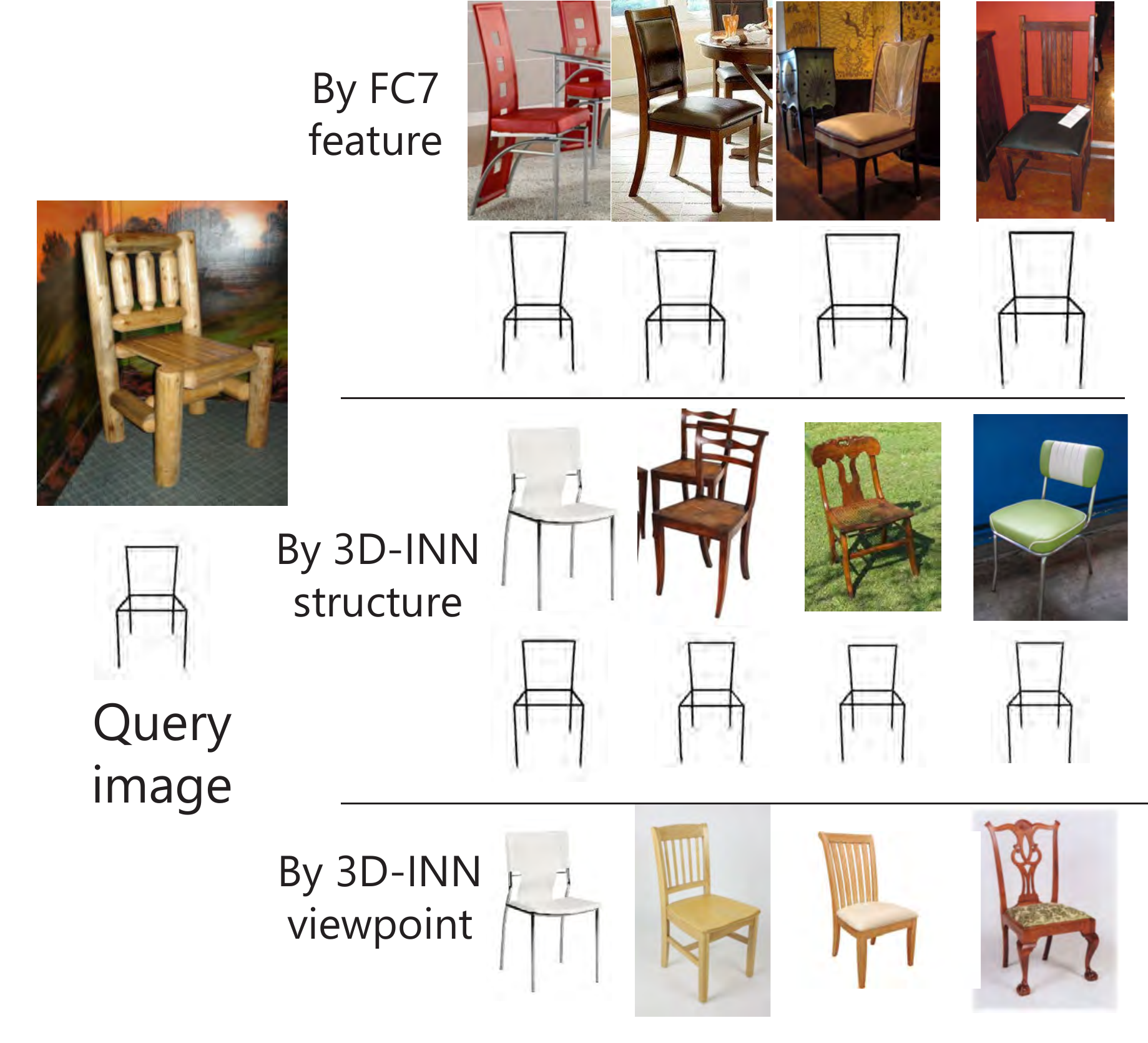}
    \caption{Retrieval results in different feature spaces. \modelshort helps to retrieve objects with similar 3D structures or similar viewpoints.} 
    \label{fig:retrieval}
    \end{minipage}
    \hfill
    \begin{minipage}{0.58\linewidth}
        \includegraphics[width=\linewidth]{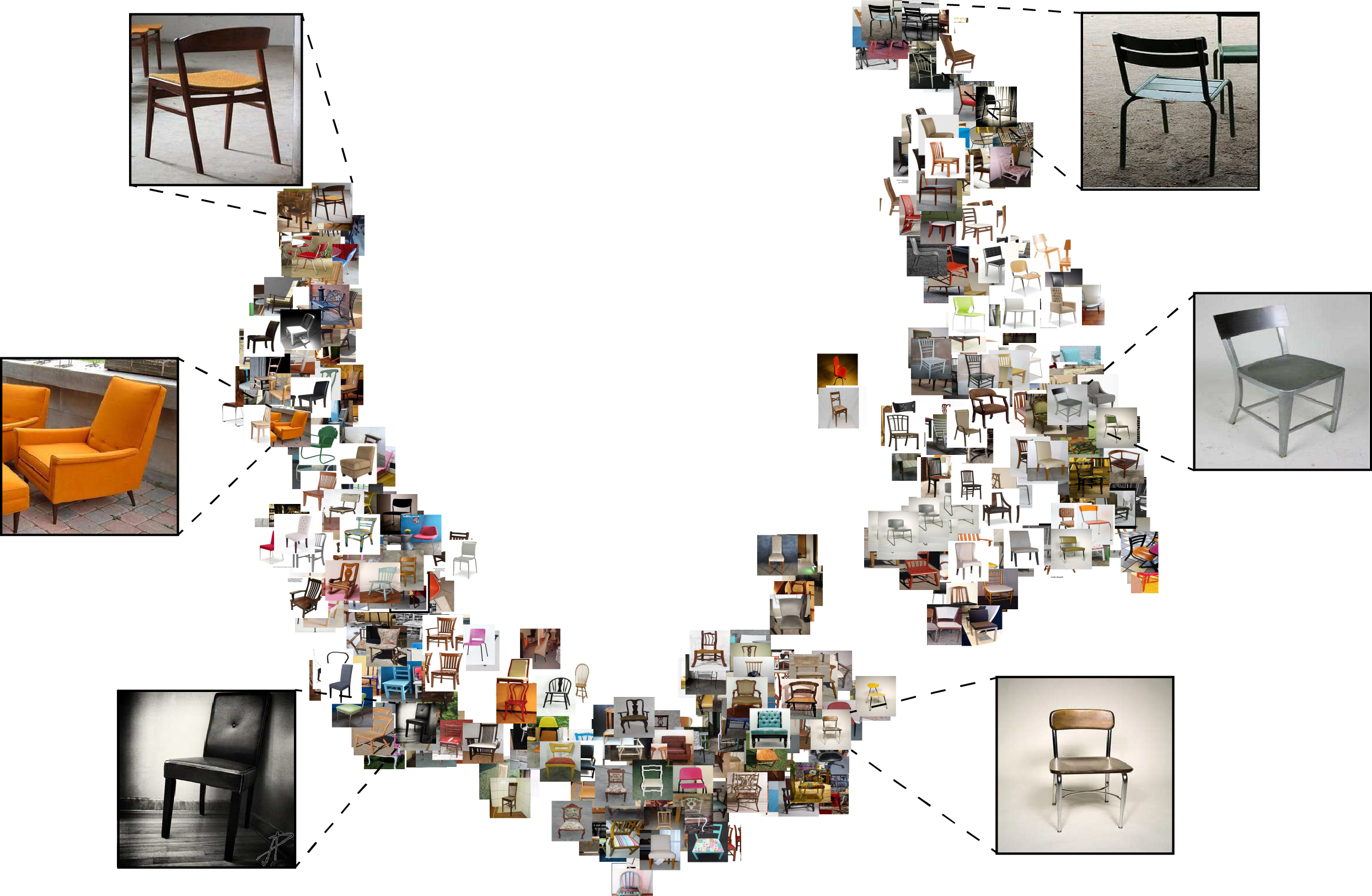}
    \caption{Object graph visualization based on learned object representations: we visualize images using t-SNE~\cite{tsne} on predicted 3D viewpoint by~\modelshort.}
    \label{fig:objgraph}
    \end{minipage}
\end{figure}

\vpar{Image Retrieval}
Using estimated 3D structural and viewpoint information, we can retrieve images based on their 3D configurations. Figure~\ref{fig:retrieval} shows image retrieval results using FC7 features from AlexNet~\cite{alexnet} and using the 3D structure and viewpoint learned by \modelshort. Our retrieval database includes all testing images of chairs and sofas in \datasetName.
In each row, we sort the best matches of the query image, measured by Euclidean distance in a specific feature space. We retrieve images in two ways: \emph{by structure}
uses estimated internal structural parameters ($\{\alpha_i\}$ in Equation~\ref{eq:pnp}), and \emph{by viewpoint} uses estimated external viewpoint parameters ($R$ in Equation~\ref{eq:pnp}).

\vpar{Object Graph}
Similar to the retrieval task, we visualize all test images for chairs in \datasetName in \fig{fig:objgraph}, using t-SNE~\cite{tsne} on estimated 3D viewpoints. Note the smooth transition from the chairs facing left to those facing right. 

\section{Conclusion}

In this paper, we introduced \model (\modelshort), which recovers the 2D keypoint and 3D structure of a (possibly deformable) object given a single image. To achieve this goal, we used 3D skeletons as an abstract 3D representation, incorporated a projection layer to the network for learning 3D parameters from 2D labels, and employed keypoint heatmaps to connect real and synthetic data. Empirically, we showed that \modelshort performs well on both 2D keypoint estimation and 3D structure and viewpoint recovery, comparable to or better than the state-of-the-arts.
Further, various applications demonstrated the potential of the skeleton representation learned by \modelshort. 

\vspace{5pt}
\noindent
{\bf Acknowledgement \;} This work is supported by NSF Robust Intelligence 1212849 and NSF Big Data 1447476 to W.F., NSF Robust Intelligence 1524817 to A.T., ONR MURI N00014-16-1-2007 to J.B.T., Shell Research, and the Center for Brain, Minds and Machines (NSF STC award CCF-1231216). The authors would like to thank Nvidia for GPU donations. Part of this work was done during Jiajun Wu's internship at Facebook AI Research.

\bibliographystyle{splncs03}
\bibliography{3dinn}

\begin{thebibliography}{10}
\providecommand{\url}[1]{\texttt{#1}}
\providecommand{\urlprefix}{URL }

\bibitem{akhter2015pose}
Akhter, I., Black, M.J.: Pose-conditioned joint angle limits for 3d human pose
  reconstruction. In: CVPR (2015)

\bibitem{Aubry14}
Aubry, M., Maturana, D., Efros, A., Russell, B., Sivic, J.: Seeing 3d chairs:
  exemplar part-based 2d-3d alignment using a large dataset of cad models. In:
  CVPR (2014)

\bibitem{bansal2016marr}
Bansal, A., Russell, B.: Marr revisited: 2d-3d alignment via surface normal
  prediction. In: CVPR (2016)

\bibitem{consensus}
Belhumeur, P.N., Jacobs, D.W., Kriegman, D.J., Kumar, N.: Localizing parts of
  faces using a consensus of exemplars. IEEE TPAMI  35(12),  2930--2940 (2013)

\bibitem{bever2010analysis}
Bever, T.G., Poeppel, D.: Analysis by synthesis: a (re-) emerging program of
  research for language and vision. Biolinguistics  4(2-3),  174--200 (2010)

\bibitem{poselet}
Bourdev, L., Maji, S., Brox, T., Malik, J.: Detecting people using mutually
  consistent poselet activations. In: ECCV (2010)

\bibitem{carreira2015human}
Carreira, J., Agrawal, P., Fragkiadaki, K., Malik, J.: Human pose estimation
  with iterative error feedback. In: CVPR (2016)

\bibitem{choy20163d}
Choy, C.B., Xu, D., Gwak, J., Chen, K., Savarese, S.: 3d-r2n2: A unified
  approach for single and multi-view 3d object reconstruction. In: ECCV (2016)

\bibitem{DB15}
Dosovitskiy, A., Tobias~Springenberg, J., Brox, T.: Learning to generate chairs
  with convolutional neural networks. In: CVPR (2015)

\bibitem{urtasun_3d_car}
Fidler, S., Dickinson, S.J., Urtasun, R.: 3d object detection and viewpoint
  estimation with a deformable 3d cuboid model. In: NIPS (2012)

\bibitem{RCNN}
Girshick, R., Donahue, J., Darrell, T., Malik, J.: Rich feature hierarchies for
  accurate object detection and semantic segmentation. In: CVPR (2014)

\bibitem{synthesis3d}
Hejrati, M., Ramanan, D.: Analysis by synthesis: 3d object recognition by
  object reconstruction. In: CVPR (2014)

\bibitem{hejrati2012analyzing}
Hejrati, M., Ramanan, D.: Analyzing 3d objects in cluttered images. In: NIPS
  (2012)

\bibitem{hinton1997generative}
Hinton, G.E., Ghahramani, Z.: Generative models for discovering sparse
  distributed representations. Philosophical Transactions of the Royal Society
  of London B: Biological Sciences  352(1358),  1177--1190 (1997)

\bibitem{hu2015learning}
Hu, W., Zhu, S.C.: Learning 3d object templates by quantizing geometry and
  appearance spaces. IEEE TPAMI  37(6),  1190--1205 (2015)

\bibitem{huang2015single}
Huang, Q., Wang, H., Koltun, V.: Single-view reconstruction via joint analysis
  of image and shape collections. ACM SIGGRAPH  34(4), ~87 (2015)

\bibitem{jaderberg2015spatial}
Jaderberg, M., Simonyan, K., Zisserman, A., Kavukcuoglu, K.: Spatial
  transformer networks. In: NIPS (2015)

\bibitem{kar2015category}
Kar, A., Tulsiani, S., Carreira, J., Malik, J.: Category-specific object
  reconstruction from a single image. In: CVPR (2015)

\bibitem{alexnet}
Krizhevsky, A., Sutskever, I., Hinton, G.E.: Imagenet classification with deep
  convolutional neural networks. In: NIPS (2012)

\bibitem{kulkarni2015picture}
Kulkarni, T.D., Kohli, P., Tenenbaum, J.B., Mansinghka, V.: Picture: A
  probabilistic programming language for scene perception. In: CVPR (2015)

\bibitem{kulkarni2015deep}
Kulkarni, T.D., Whitney, W.F., Kohli, P., Tenenbaum, J.B.: Deep convolutional
  inverse graphics network. In: NIPS (2015)

\bibitem{leclerc1992optimization}
Leclerc, Y.G., Fischler, M.A.: An optimization-based approach to the
  interpretation of single line drawings as 3d wire frames. IJCV  9(2),
  113--136 (1992)

\bibitem{li2015joint}
Li, Y., Su, H., Qi, C.R., Fish, N., Cohen-Or, D., Guibas, L.J.: Joint
  embeddings of shapes and images via cnn image purification. ACM SIGGRAPH Asia
   34(6),  234 (2015)

\bibitem{fpm}
Lim, J.J., Khosla, A., Torralba, A.: {FPM}: Fine pose parts-based model with 3d
  cad models. In: ECCV (2014)

\bibitem{ikea}
Lim, J.J., Pirsiavash, H., Torralba, A.: Parsing ikea objects: Fine pose
  estimation. In: ICCV (2013)

\bibitem{liu2013bird}
Liu, J., Belhumeur, P.N.: Bird part localization using exemplar-based models
  with enforced pose and subcategory consistency. In: ICCV (2013)

\bibitem{lowe1987three}
Lowe, D.G.: Three-dimensional object recognition from single two-dimensional
  images. Artificial intelligence  31(3),  355--395 (1987)

\bibitem{tsne}
Van~der Maaten, L., Hinton, G.: Visualizing data using t-sne. JMLR  9(11),
  2579--2605 (2008)

\bibitem{newell2016stacked}
Newell, A., Yang, K., Deng, J.: Stacked hourglass networks for human pose
  estimation. In: ECCV (2016)

\bibitem{peng2014exploring}
Peng, X., Sun, B., Ali, K., Saenko, K.: Exploring invariances in deep
  convolutional neural networks using synthetic images. CoRR, abs/1412.7122  2
  (2014)

\bibitem{pepik2012teaching}
Pepik, B., Stark, M., Gehler, P., Schiele, B.: Teaching 3d geometry to
  deformable part models. In: CVPR (2012)

\bibitem{prasad2010finding}
Prasad, M., Fitzgibbon, A., Zisserman, A., Van~Gool, L.: Finding nemo:
  Deformable object class modelling using curve matching. In: CVPR (2010)

\bibitem{ramakrishna2012reconstructing}
Ramakrishna, V., Kanade, T., Sheikh, Y.: Reconstructing 3d human pose from 2d
  image landmarks. In: ECCV (2012)

\bibitem{FastRCNN}
Ren, S., He, K., Girshick, R., Sun, J.: Faster {R-CNN}: Towards real-time
  object detection with region proposal networks. In: NIPS (2015)

\bibitem{FLIC}
Sapp, B., Taskar, B.: Modec: Multimodal decomposable models for human pose
  estimation. In: CVPR (2013)

\bibitem{satkin_bmvc2012}
Satkin, S., Lin, J., Hebert, M.: Data-driven scene understanding from 3{D}
  models. In: BMVC (2012)

\bibitem{shakhnarovich2003fast}
Shakhnarovich, G., Viola, P., Darrell, T.: Fast pose estimation with
  parameter-sensitive hashing. In: ICCV (2003)

\bibitem{shih2015part}
Shih, K.J., Mallya, A., Singh, S., Hoiem, D.: Part localization using
  multi-proposal consensus for fine-grained categorization. In: BMVC (2015)

\bibitem{shrivastava2013building}
Shrivastava, A., Gupta, A.: Building part-based object detectors via 3d
  geometry. In: ICCV (2013)

\bibitem{haosu_sig14}
Su, H., Huang, Q., Mitra, N.J., Li, Y., Guibas, L.: Estimating image depth
  using shape collections. ACM TOG  33(4), ~37 (2014)

\bibitem{su15}
Su, H., Qi, C.R., Li, Y., Guibas, L.: Render for cnn: Viewpoint estimation in
  images using cnns trained with rendered 3d model views. In: ICCV (2015)

\bibitem{sun2014virtual}
Sun, B., Saenko, K.: From virtual to reality: Fast adaptation of virtual object
  detectors to real domains. In: BMVC (2014)

\bibitem{deepface2}
Taigman, Y., Yang, M., Ranzato, M., Wolf, L.: Web-scale training for face
  identification. In: CVPR (2015)

\bibitem{tompson2015efficient}
Tompson, J., Goroshin, R., Jain, A., LeCun, Y., Bregler, C.: Efficient object
  localization using convolutional networks. In: CVPR (2015)

\bibitem{tompson2014joint}
Tompson, J.J., Jain, A., LeCun, Y., Bregler, C.: Joint training of a
  convolutional network and a graphical model for human pose estimation. In:
  NIPS (2014)

\bibitem{datasetbias}
Torralba, A., Efros, A.A.: Unbiased look at dataset bias. In: CVPR (2011)

\bibitem{non_rigid3d}
Torresani, L., Hertzmann, A., Bregler, C.: Learning non-rigid 3d shape from 2d
  motion. In: NIPS (2003)

\bibitem{toshev2014deeppose}
Toshev, A., Szegedy, C.: Deeppose: Human pose estimation via deep neural
  networks. In: CVPR. pp. 1653--1660 (2014)

\bibitem{tulsiani2015viewpoints}
Tulsiani, S., Malik, J.: Viewpoints and keypoints. In: CVPR (2015)

\bibitem{vicente2014reconstructing}
Vicente, S., Carreira, J., Agapito, L., Batista, J.: Reconstructing pascal voc.
  In: CVPR (2014)

\bibitem{CUB}
Wah, C., Branson, S., Welinder, P., Perona, P., Belongie, S.: {The Caltech-UCSD
  Birds-200-2011 Dataset}. Tech. Rep. CNS-TR-2011-001, California Institute of
  Technology (2011)

\bibitem{wu2015galileo}
Wu, J., Yildirim, I., Lim, J.J., Freeman, B., Tenenbaum, J.: Galileo:
  Perceiving physical object properties by integrating a physics engine with
  deep learning. In: NIPS (2015)

\bibitem{pascal3d}
Xiang, Y., Mottaghi, R., Savarese, S.: Beyond pascal: A benchmark for 3d object
  detection in the wild. In: WACV (2014)

\bibitem{SUN}
Xiao, J., Hays, J., Ehinger, K., Oliva, A., Torralba, A.: Sun database:
  Large-scale scene recognition from abbey to zoo. In: CVPR (2010)

\bibitem{xue2012example}
Xue, T., Liu, J., Tang, X.: Example-based 3d object reconstruction from line
  drawings. In: CVPR (2012)

\bibitem{yang2011articulated}
Yang, Y., Ramanan, D.: Articulated pose estimation with flexible
  mixtures-of-parts. In: CVPR (2011)

\bibitem{yasin2016dualsource}
Yasin, H., Iqbal, U., Kr{\"u}ger, B., Weber, A., Gall, J.: A dual-source
  approach for 3d pose estimation from a single image. In: CVPR (2016)

\bibitem{yuille2006vision}
Yuille, A., Kersten, D.: Vision as bayesian inference: analysis by synthesis?
  Trends in cognitive sciences  10(7),  301--308 (2006)

\bibitem{zeng20163dmatch}
Zeng, A., Song, S., Nie{\ss}ner, M., Fisher, M., Xiao, J.: 3dmatch: Learning
  the matching of local 3d geometry in range scans. arXiv preprint
  arXiv:1603.08182  (2016)

\bibitem{zhou2016learning}
Zhou, T., Kr{\"a}henb{\"u}hl, P., Aubry, M., Huang, Q., Efros, A.A.: Learning
  dense correspondence via 3d-guided cycle consistency. In: CVPR (2016)

\bibitem{zhou153d}
Zhou, X., Leonardos, S., Hu, X., Daniilidis, K.: 3d shape reconstruction from
  2d landmarks: A convex formulation. In: CVPR (2015)

\bibitem{zia2013detailed}
Zia, M.Z., Stark, M., Schiele, B., Schindler, K.: Detailed 3d representations
  for object recognition and modeling. IEEE TPAMI  35(11),  2608--2623 (2013)

\end{thebibliography}
\end{document}